%% file: main.tex
\documentclass[10pt,twocolumn,letterpaper]{article}

 \usepackage{cvpr}              %

\usepackage[accsupp]{axessibility}  %
\input{preamble}

\definecolor{cvprblue}{rgb}{0.21,0.49,0.74}
\usepackage[pagebackref,breaklinks,colorlinks,allcolors=cvprblue]{hyperref}

\def\paperID{15598} %
\def\confName{CVPR}
\def\confYear{2025}

\title{VladVA: Discriminative Fine-tuning of LVLMs}

\author{%
Yassine Ouali*$^1$ \quad  Adrian Bulat*$^{1,2}$ \quad Alexandros Xenos$^{1,3}$ \quad Anestis Zaganidis$^1$ \\ Ioannis Maniadis Metaxas$^1$ \quad Brais Martinez$^1$ \quad Georgios Tzimiropoulos$^{1,3}$\\
$^1$Samsung AI Cambridge \quad $^2$Technical University of Iasi \quad $^3$Queen Mary University of London
}

\begin{document}
\maketitle
\input{sec/0_abstract}    
\input{sec/1_intro}

\input{sec/2_related_work}

\input{sec/3_method}

\input{sec/4_results}

\input{sec/5_ablation_and_analysis}

\input{sec/6_conclusions}
 \input{sec/X_suppl}
\clearpage
{
    \small
    \bibliographystyle{ieeenat_fullname}
    \bibliography{main}
}

\end{document}

%% file: preamble.tex
\usepackage[utf8]{inputenc}

\newcommand{\ITT}{ITT\xspace}
\newcommand{\TOT}{TOT\xspace}

\usepackage{adjustbox}
\usepackage{csquotes}
\usepackage{epigraph}
\usepackage{tcolorbox}
\usepackage{colortbl}

\usepackage{listings}
\usepackage{xcolor}

\usepackage{multicol}
\usepackage{multirow}

\lstdefinestyle{pythonstyle}{
    language=Python,
    basicstyle=\ttfamily\footnotesize,
    keywordstyle=\color{blue},
    stringstyle=\color{green},
    commentstyle=\color{gray},
    morecomment=[l][\color{magenta}]{\#}
}

\definecolor{mygreen}{HTML}{3cb44b}
\definecolor{myred}{HTML}{800020}
\definecolor{myorange}{HTML}{FFBF00}
\newcommand{\PredSty}[1]{\textnormal{\ttfamily\color{mygreen!90!black}#1}\unskip}
\newcommand{\PredStyRed}[1]{\textnormal{\ttfamily\color{myred!90!black}#1}\unskip}
\newcommand{\PredStyOrange}[1]{\textnormal{\ttfamily\color{myorange!90!black}#1}\unskip}

\definecolor{textcolor}{HTML}{049e88}
\definecolor{imagecolor}{HTML}{4c71c2}

\definecolor{mygray}{rgb}{0.89, 0.93, 0.85}
\definecolor{whitesmoke}{rgb}{0.96, 0.96, 0.96}
\definecolor{timberwolf}{rgb}{0.86, 0.84, 0.82}
\definecolor{darkgreen}{rgb}{0.0, 0.5, 0.0}
\definecolor{lightgray}{rgb}{0.9, 0.9, 0.9}
\definecolor{Mycolor1}{HTML}{BAD8F2}
\definecolor{Mycolor2}{HTML}{E0F0FA}

\newcolumntype{a}{>{\columncolor{mygray}}c}

\usepackage{amssymb}%
\usepackage{pifont}%
\newcommand{\xmark}{\ding{55}}%

\makeatletter
\g@addto@macro{\endtabular}{\rowfont{}}%
\makeatother
\newcommand{\rowfonttype}{}%
\newcommand{\rowfont}[1]{%
\gdef\rowfonttype{#1}#1\ignorespaces%
}
\makeatother

\input{math_commands.tex}

\usepackage{stfloats}
\usepackage{float}

%% file: math_commands.tex
\usepackage{amsmath,amsfonts,bm}

\def\eqref#1{equation~\ref{#1}}

\def\1{\bm{1}}

\def\vx{{\bm{x}}}

\DeclareMathAlphabet{\mathsfit}{\encodingdefault}{\sfdefault}{m}{sl}
\SetMathAlphabet{\mathsfit}{bold}{\encodingdefault}{\sfdefault}{bx}{n}

\def\gL{{\mathcal{L}}}

\def \xi{\vx^I}

%% file: sec/0_abstract.tex
\begin{abstract}
Contrastively-trained Vision-Language Models (VLMs) like CLIP have become the de facto approach for discriminative vision-language representation learning. However, these models have limited language understanding, often exhibiting a ``bag of words'' behavior. At the same time, Large Vision-Language Models (LVLMs), which combine vision encoders with LLMs, have been shown to be capable of detailed vision-language reasoning, yet their autoregressive nature renders them less suitable for discriminative tasks.

In this work, we propose to combine ``the best of both worlds'': a new training approach for discriminative fine-tuning of LVLMs that results in strong discriminative and compositional capabilities. Essentially, our approach converts a \textit{generative} LVLM into a \textit{discriminative} one, unlocking its capability for powerful image-text discrimination combined with enhanced language understanding.

Our contributions include (1) A carefully designed training/optimization framework that utilizes image-text pairs of variable length and granularity for training the model with both contrastive and next-token prediction losses. This is accompanied by ablation studies that justify the necessity of our framework's components. (2) A parameter-efficient adaptation method using a combination of soft prompting and LoRA adapters. (3) Significant improvements over state-of-the-art CLIP-like models of similar size, including standard image-text retrieval benchmarks and notable gains in compositionality.

\end{abstract}

%% file: sec/1_intro.tex
\section{Introduction}
\label{sec:intro}

Contrastively-trained Vision Language Models (VLMs) (e.g. CLIP~\cite{radford2021learning}) have become the predominant direction for vision-language representation learning, exhibiting remarkable zero-shot abilities~\cite{radford2021learning,zhai2023sigmoid,jia2021scaling,li2022blip,li2021supervision}. However, the great success of these models in many vision-language and vision tasks, even in a zero-shot manner, ``sweeps under the rug'' some of their important limitations. Specifically, such models struggle to exhibit advanced language understanding capabilities, suffer from a limited understanding of compositionality, and manifest a \textit{bag of words} behavior~\cite{lewis2022does,yuksekgonul2022and}. For example, even with \textit{bag of words} behavior, VLMs have shown remarkable zero-shot retrieval accuracy on the Flickr~\cite{young2014image} and COCO~\cite{lin2014microsoft} datasets. Still, they perform poorly on a simple word order permutation task on the same datasets~\cite{yuksekgonul2022and}. Unfortunately, these issues persist even when the model and the dataset size increase~\cite{hsieh2024sugarcrepe}.

Concomitantly, inspired by the success of LLMs~\cite{brown2020language,touvron2023llama} in acting as generalist assistants~\cite{dubey2024llama}, a series of works combine pretrained vision encoders and LLMs~\cite{li2024llavanext,li2024llava,zhu2023minigpt} to construct Large Vision-Language Models (LVLMs) capable of performing interactive multi-modal 
conversations. Among others, these models have been shown capable of \textit{exhibiting strong reasoning and vision-language understanding capabilities}, offering fine-grained and detailed responses~\cite{li2024llava,li2024llavanext,chen2023measuring,deitke2024molmo}. However, they are trained with a next-token prediction loss in an autoregressive manner, which appears less suitable for direct utilization in discriminative image-text tasks (\eg image-text retrieval). 

To our knowledge, the very recent (concurrent) work ~\cite{jiang2024e5} is the first one to show that, with appropriate prompting, LVLMs can serve as zero-shot discriminative models. Importantly,~\cite{jiang2024e5} advocates for a text-text optimization approach, stating that contrastive image-text fine-tuning has a detrimental effect on the model's performance. In contrast to~\cite{jiang2024e5}, we propose a new training framework for discriminative image-text fine-tuning of LVLMs, aiming to convert the original \textit{generative} LVLM into a \textit{discriminative} one, thereby significantly enhancing its capability for image-text discrimination while preserving the compositional strengths of the original model.

In our approach, following the (independent) two-towers paradigm, the vision embeddings are produced by passing the image through the entire LVLM, and the text embeddings by passing the text through the LLM of the LVLM. Intuitively, for the vision embedding, the LLM acts as an information processor that refines the visual information while simultaneously aligning it with the textual representations. We coin our approach \textbf{VladVA}: \textbf{V}ision-\textbf{L}anguage \textbf{A}daptation for \textbf{D}iscriminative \textbf{V}isual \textbf{A}ssistant. Our \textbf{main contributions} are: 
\begin{itemize}
\item
We devise a carefully designed optimization framework that utilizes image-text pairs of variable length and granularity for model training (\ie both short and long captions). Using this data, the model is trained with both contrastive and next-token prediction losses, which are both shown to be necessary for unlocking strong discrimination and compositionality capabilities. Our design choices are accompanied by ablation studies, which justify the necessity of our framework's components. 
\item 
To facilitate efficient training, we show how the model can be fine-tuned using a parameter-efficient adaptation method based on a combination of soft prompting~\cite{li2021prefix} and LoRA adapters ~\cite{hu2021lora}. We show the positive impact of both components. 
\item 
We report significant improvements over state-of-the-art two-tower models (e.g. CLIP-like models) of similar size on standard image-text retrieval benchmarks (+4.7-7.0\% gains in absolute terms). Moreover, we report notable gains on several vision-language understanding and compositionality benchmarks (up to +15\%).
\end{itemize}

%% file: sec/2_related_work.tex
\section{Related work}
\label{sec:related_work}

\subsection{Large Vision Language Models (LVLMs)}

Inspired by breakthrough research in language modeling~\cite{brown2020language,touvron2023llama,jiang2023mistral,team2024gemma}, a series of methods seek to combine pretrained LLMs and vision encoders to construct Large Vision Language Models (LVLMs) capable of processing image-text data jointly~\cite{liu2024visual,liu2024improved,wu2023visual,zhu2023minigpt,wang2023cogvlm,bai2023qwen,li2024mini,deitke2024molmo}. The prevalent strategy consists in aligning the features produced by a pretrained vision encoder to the textual space assumed by a pretrained LLM using a projection module, \eg LLaVA~\cite{liu2024visual}, following a two-stage alignment procedure. Follow-up works expand this to interleaved image-text data~\cite{li2024llava,abdin2024phi} and multiple input crops~\cite{abdin2024phi} while seeking to improve the model's efficiency~\cite{chu2024mobilevlm}.

Despite their strong generative and comprehension abilities~\cite{liu2024improved}, current LVLMs are primarily restricted to generative tasks. Only very recently, Jiang \etal~\cite{jiang2024e5}, inspired by the recent progress in NLP~\cite{bitton2023visit,lee2024nv} adapted a LLaVA-NeXT~\cite{li2024llavanext} model to discriminative tasks using a contrastive-like loss and text data only. We note that unlike~\cite{jiang2024e5}, we introduce a training framework that learns from multi-turn image-text pairs (as opposed to text only) using a novel formulation that jointly combines a contrastive loss with a next-token prediction, reflecting the data characteristics and inducing a gradual representation buildup. Concurrently, VLM2Vec~\cite{jiang2024vlm2vec} adapts an LVLM for multi-modal retrieval. However, it uses a different loss and training strategy (no generative loss, no short-long captions training, no soft prompting). We compare our approach with both E5-V and VLM2Vec, significantly improving upon their results despite using smaller/lighter models.

\subsection{Discriminative Vision-Language Models}

The prevalent approach for training Discriminative VLMs follows the two-tower contrastive approach pioneered by CLIP~\cite{radford2021learning}, whereby an image and text encoder are trained on web-collected image-text pairs to learn a joint multi-modal (\ie vision and language) space. Subsequent works build upon CLIP by scaling the data~\cite{zhai2023sigmoid,schuhmann2022laion,bulat2024efficient}, improving the architecture using late/early interactions~\cite{li2023blip} or improving the training loss~\cite{zhai2023sigmoid,bulat2024fff}. Despite their remarkable zero-shot and representation learning abilities~\cite{radford2021learning} such models were shown to have significant shortcomings related to limited language understanding capabilities, including: lack of compositionality understanding~\cite{lewis2022does}, manifesting \textit{bag of words} behavior~\cite{yuksekgonul2022and}, struggling with spatial relations~\cite{lewis2022does}, being susceptible to typographical attacks~\cite{goh2021multimodal}, \etc. Recent works aim to address these shortcomings by constructing synthetic hard negatives~\cite{yuksekgonul2022and} or performing cross-modality attention~\cite{li2023blip}. However, the former does not inherently change the model's behaviors and has been shown to potentially learn a series of shortcuts/artifacts~\cite{hsieh2024sugarcrepe}. Meanwhile, the latter is impractical for deployment at scale, as, due to the interactions between the encoders, each new query incurs an additional inference for every image within the set.

To alleviate these shortcomings and improve the overall capabilities of such models, we depart from the prevalent approach of training VLMs using a contrastive loss and, instead, propose a new approach that seeks to convert generative LVLMs into discriminative models by adapting them using a newly proposed framework that combines generative and discriminative objectives.

%% file: sec/3_method.tex
\section{Method}
\label{sec:method}

Herein, we present VladVA (\textbf{V}ision-\textbf{L}anguage \textbf{A}daptation for \textbf{D}iscriminative \textbf{V}isual \textbf{A}ssistant), our novel approach for discriminative fine-tuning of LVLMs that results in strong discriminative and compositional capabilities. This section is structured as follows: Sec.~\ref{ssec:background} briefly introduces the architecture, detailing how LVLMs can be used as discriminators in a zero-shot manner. Sec.~\ref{ssec:method-generation-to-discrimination} details the core component of our approach: a carefully designed optimization framework that utilizes image-text pairs of variable length and granularity for training the model with both contrastive (Sec.~\ref{ssec:method-c})  and next-token prediction (Sec.~\ref{ssec:method-ar}) losses, showcasing that contrastive is best with short captions and autoregressive with long captions. In Sec.~\ref{ssec:method-efficient}, we present our parameter-efficient adaptation, while Sec.~\ref{ssec:model_beh} analyzes how the model's behavior changes after training.

\begin{figure*}[!ht]
    \centering
    \includegraphics[trim={0.5cm 0.25cm 0.5cm 0},clip,width=0.7\linewidth]{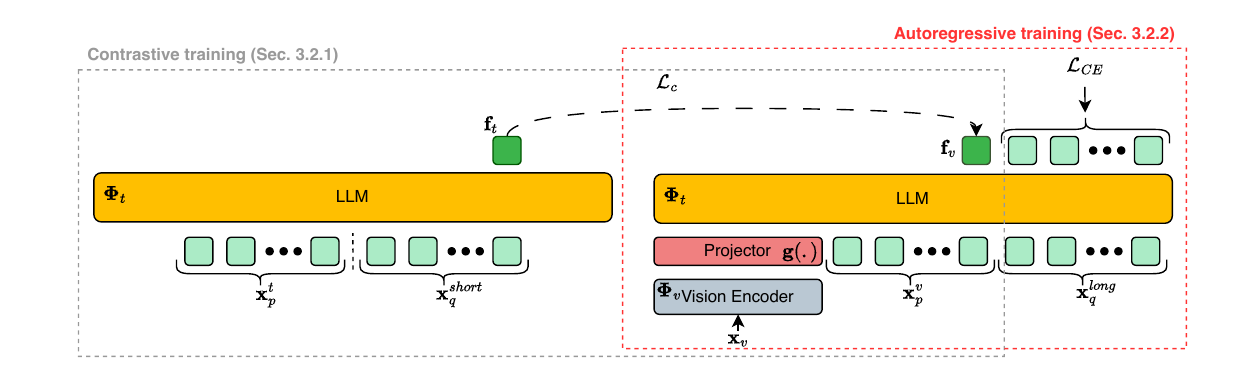}
        \vspace*{-0.3cm}
    \caption{\textbf{Overall VladVA framework:} a generative LVLM is adapted into a discriminative model with the help of (1) a contrastive training loss (Sec.~\ref{ssec:method-c}), and (2) an autoregressive loss (Sec.~\ref{ssec:method-ar}). The first one is applied on image-text pairs with short(er) captions, encouraging the last token produced by both modalities to be discriminative. The second one, jointly optimized with the first one, is applied only on longer captions and allows the model to learn fine-grained details.
    }
    \label{fig:overall_framework}
    \vspace*{-0.3cm}
\end{figure*}

\subsection{LVLMs as zero-shot discriminative models}~\label{ssec:background}
LVLMs consist of an LLM $\Phi_t$, a vision encoder $\Phi_v$, and a module $g$ that projects the vision features into the LLM's textual space. Once fine-tuned, such models can produce a textual answer $\mathbf{x}_a = \Phi_t(g(\Phi_v(\mathbf{x}_v)),\mathbf{x}_q)$ when presented with an input image $\mathbf{x}_{v}$ and a text query (or prompt) $\mathbf{x}_{q}$. 

Despite being solely trained with an autoregressive next-token prediction loss on limited amounts of data ($<5$M), such models can act as multi-modal discriminative models in a zero-shot manner~\cite{jiang2024e5}. To elicit this capability, the image embedding $\mathbf{f}_{v}=\Phi_t(g(\Phi_v(\mathbf{x}_v),\mathbf{x}^v_p))[eos]$ is obtained by passing the image alongside a handcrafted image prompt $\mathbf{x}^v_p$ (e.g., ``in one word, describe the image'') through the LVLM and taking the output representation of the last token. Analogously, the text embedding $\mathbf{f}_{t}=\Phi_t(\mathbf{x}^t_p,\mathbf{x}_q)[eos]$ is produced by passing the handcrafted text prompt $\mathbf{x}^t_p$ (e.g., ``in one word, describe the text'') and input query $\mathbf{x}_q$ through the LLM (of the LVLM) and taking again the output representation of the last token. We will refer to these particular tokens as ``summary tokens'' (summarizing image and text information, respectively).
Note that, typically, the respective handcrafted prompts for the image ($\mathbf{x}^v_p$) and text ($\mathbf{x}^t_p$) modalities are different. Finally, the similarity between an image and a text query can be computed by taking the cosine similarity between the two: $s = \texttt{cos\_sim}(\mathbf{f}_v, \mathbf{f}_t)$.

\begin{figure}[!ht]
    \centering
    \includegraphics[trim={0cm 0cm 0cm 1cm},clip,width=1.0\linewidth]{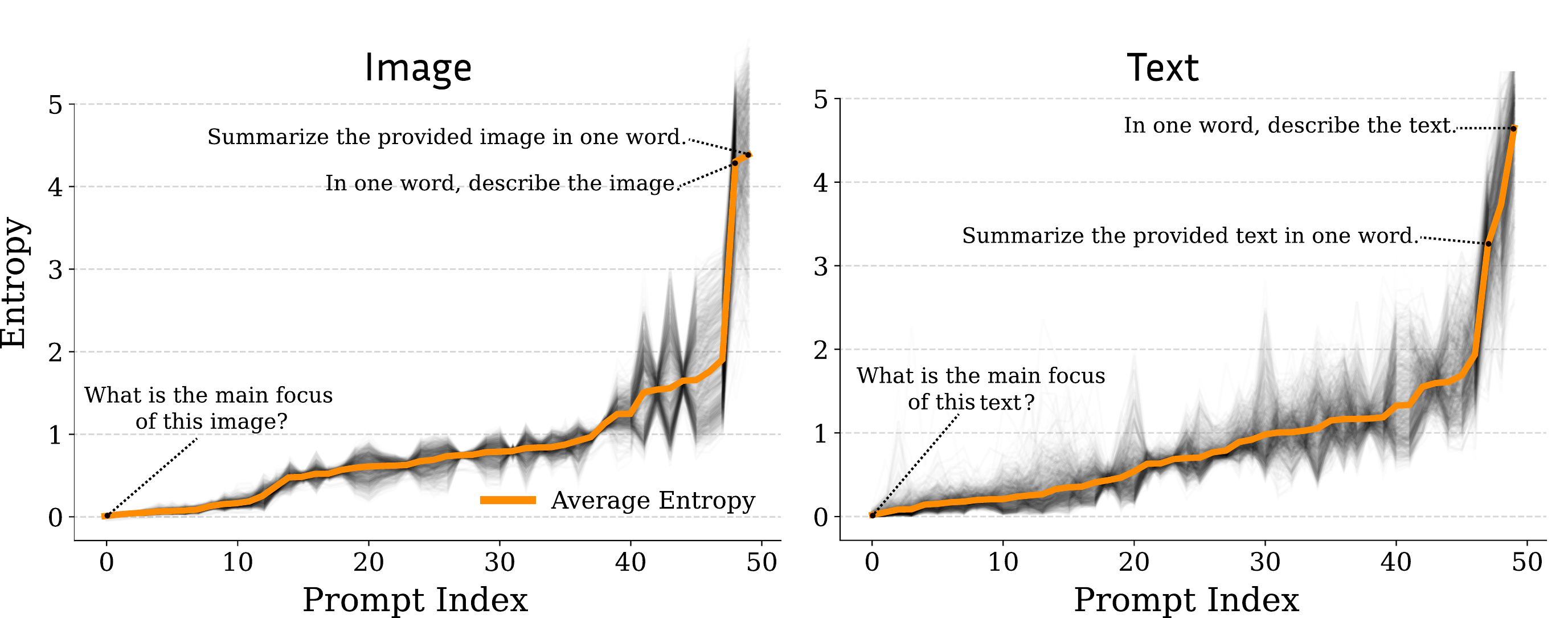}
    \caption{\textbf{Entropy of the output probability distribution} at the next-to-be-predicted token location using a LLaVA-1.5-7B
    for a set of 50 prompts for both images and captions.}
    \label{fig:entropy}
    \vspace{-0.3cm}
\end{figure}

\begin{figure}[!ht]
    \centering
    \includegraphics[trim={0cm 0cm 0cm 0cm},clip,width=1.0\linewidth]{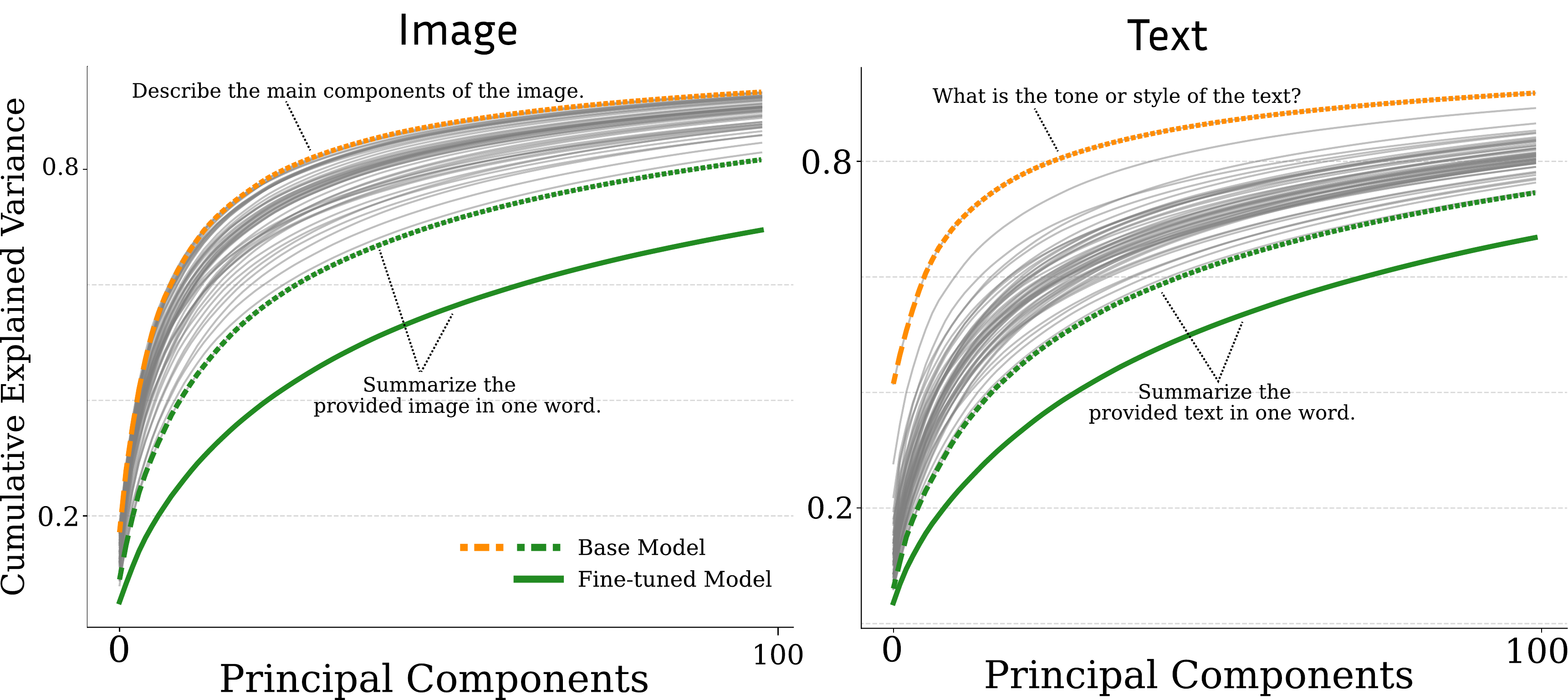}
    \caption{\textbf{Cumulative variance of the image and text embedding matrices}
     over a set of 50 prompts on Flickr30k.
     Embeddings that capture more information about the input translate
     into a cumulative variance that requires more principal components to be explained, \ie a higher-rank embedding matrix.}
    \label{fig:cumvar}
    \vspace{-0.6cm}
\end{figure}

\begin{figure*}[!ht]
    \centering
    \includegraphics[width=1.0\textwidth]{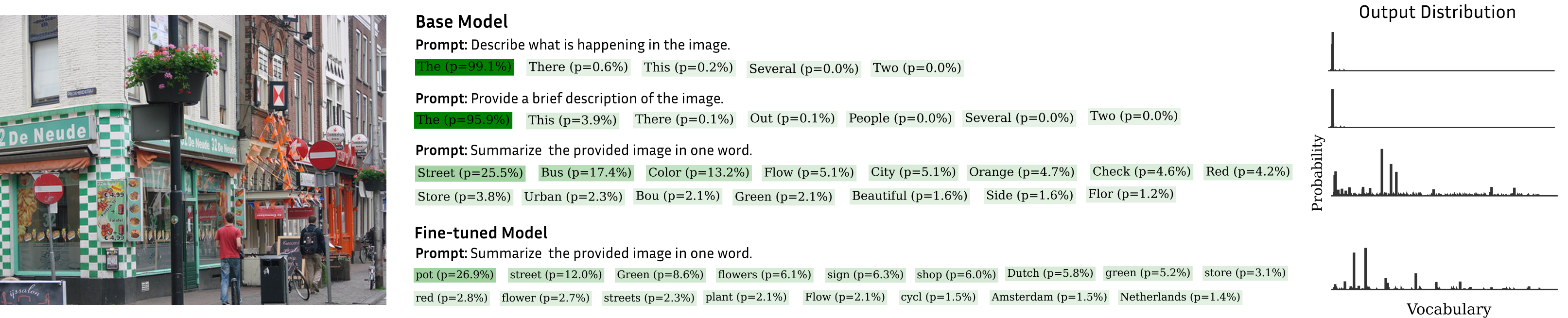}
    \caption{\textbf{Top-k next-to-be-predicted tokens} before and after VladVA fine-tuning (our approach). On the right, we show the output probability distribution for each case.
    When using the best prompt (``Summarize the provided image in one word''), the representations of the next token can encode diverse
    and more discriminative information, making potentially better-quality embeddings. This behavior
    is further improved after VladVA fine-tuning.}
    \label{fig:decoded-tokens}
    \vspace*{-0.25cm}
\end{figure*}

\begin{figure}
    \centering
    \includegraphics[width=0.6\linewidth]{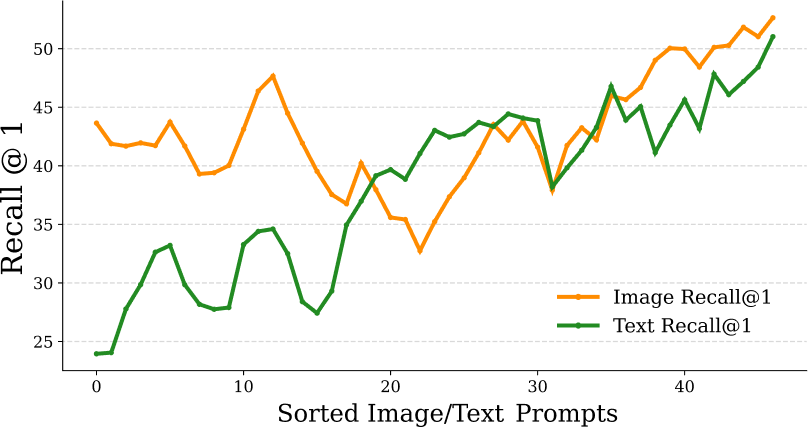}
    \caption{\textbf{Image and text retrieval score on Flickr30k} over a
    set of 50 image-text
    prompts ordered by their entropy scores (Fig.~\ref{fig:entropy}).
    We can observe that prompts with high average entropy scores
    correlate positively with the zero-shot retrieval performance.}
    \label{fig:retrieval-prompts-correlation}
    \vspace{-0.4cm}
\end{figure}

\vspace{0.1in}\noindent\textbf{What makes a good prompt?}
 Zero-shot adaptation by prompting already provides decent results despite the task changing from generation to discrimination. To shed some light, herein, we study (a) what makes a good prompt and (b) how we can identify it.

To answer these questions, we construct a testbed consisting of 1,000 image-caption pairs from Flickr30k~\cite{young2014image}, which we then use to evaluate the quality of various prompts. The prompts (50 image-text pairs in total) are constructed using ChatGPT. Each prompt pair is fed, alongside an image and its respective caption, through the LLaVA-1.5-7B model. For each image-prompt pair and caption-prompt pair, we extract the token embedding at the output position and the corresponding output probability distribution over the vocabulary. These are then used to compute two metrics for each prompt: the average entropy of its output distributions and the cumulative variance of its embeddings.
As Figs.~\ref{fig:entropy} and~\ref{fig:decoded-tokens} show, when the model is prompted with sentences
consisting of specific keywords, such as \texttt{in a few words} or in \texttt{one word}, the model
is pushed to condense the information of the image or text in the next token, resulting 
in an output distribution with high entropy. More importantly, when
investigating the generated embeddings, we observe that higher entropy prompts
result in embeddings with more spread-out cumulative variance, \ie requiring more principal components to capture the same amount of variance, indicating an embedding matrix with
a high rank (see Fig.~\ref{fig:cumvar}). This translates into discriminative embeddings that can capture more
information about the inputs, making them suitable for embedding tasks.
The benefit of this behavior is illustrated in Fig.~\ref{fig:retrieval-prompts-correlation}, which shows a positive correlation between prompts with high entropy scores and the model's zero-shot retrieval performance.
Hence, our approach should seek to produce embeddings with a) spread-out variance and b) probability distributions over the vocabulary with increased entropy. 

\subsection{Discriminative fine-tuning of LVLMs: from generation to discrimination}
\label{ssec:method-generation-to-discrimination}

Despite exhibiting surprising innate zero-shot abilities, LVLM's direct discriminative performance lags behind that of state-of-the-art contrastively trained VLMs. Hence, carefully designed frameworks are needed to unlock the full potential of such models. This is the very goal of our work: to introduce a well-grounded adaptation/training framework that surfaces the discriminative image-text capabilities of a generative LVLM.

Notably, our findings contradict those of the very recent work of~\cite{jiang2024e5}, which found that contrastive image-text fine-tuning is detrimental and limits training to \text{text-text} contrastive learning alone. This highlights the importance of our proposed approach, which overcomes such impediments and significantly boosts the discriminative performance of the model.

Having established the architecture in the previous section, the two other pillars are the data and training strategy. 

\noindent \textbf{Data strategy:} We argue for the importance of data diversity in terms of granularity and group captions according to their length: short captions ($<30$ tokens) and long captions ($30-500$ tokens). The short captions capture coarse details and summarize image content teaching the model to discriminate with regard to high-level image information. Longer captions capture finer image details and promote a better understanding of language concepts such as spatial relationships and compositionality. For a strong discriminative model, both are necessary. Therefore, for images missing either caption type, we use a BLIP2~\cite{li2023blip} captioner to generate short captions and ShareGPT-4V~\cite{chen2023sharegpt4v} to generate long captions. This allows us to leverage both supervisory signals for training.

\noindent \textbf{Training strategy:} As we demonstrate in this work, the variable length of the training data poses its own challenges: unlike the case of short captions, where training using the well-studied contrastive loss performs well, it collapses for longer captions. This brings us to the proposed training strategy, whereby, to address this challenge, we propose a hybrid training approach that combines a contrastive loss (see Sec.~\ref{ssec:method-c}) and a \textit{next-token prediction loss for discriminative adaptation} (see Sec.~\ref{ssec:method-ar}). Finally, as full model fine-tuning is computationally expensive, in Sec.~\ref{ssec:method-efficient}, we detail a fine-tuning strategy that combines adapters with soft prompting.

\subsubsection{Image-text contrastive alignment}~\label{ssec:method-c}

Under a multi-modal contrastive formulation, the image and text representations, $\mathbf{f}_v$ and $\mathbf{f}_t$ respectively, must be close if they are semantically similar and far apart otherwise, under a specified distance metric. At train time, this is enforced using a symmetric image-text and text-image contrastive loss, which, for a given mini-batch containing $b$ randomly selected samples, can be described as:
\begin{equation}
    \gL_{c} = \frac{1}{b}\sum_{k=1}^b (-\log \frac{\exp(s^{k,k}_v)}{\sum_{j} \exp(s^{k,j}_v)} - \log \frac{\exp(s^{k,k}_t)}{\sum_{j} \exp(s^{j,k}_t)}),
\end{equation}
where $s^{k,j}_v=\texttt{cos\_sim}(\mathbf{f}_v^k, \mathbf{f}_t^j)$ denotes the cosine similarity between the $k$-th image and the $j$-th caption (image-to-text), and similarity, $s^{k,j}_t$ the text-to-image similarity.

During training, the contrastive loss is applied to the very same tokens used for the zero-shot evaluation, as they represent the optimal starting point for further fine-tuning (Sec.~\ref{ssec:background}). We note that the contrastive loss is mostly suitable for training using short captions $\mathbf{x}^{short}_q$ (\ie $<30$ tokens), like the ones typically used for CLIP pre-training. We found that training the model using a contrastive loss on longer captions proves challenging. Hence, to address this, in the following section, we study and propose a new formulation that enables discriminative training on variable-length data.

\subsubsection{Autoregressive training for learning discriminative LVLM representations}
\label{ssec:method-ar}
Until now, the modality-specific embeddings are obtained by taking the last token, prior to any generation, while the training is largely focused on short (\ie $<30$ tokens) captions, mimicking the CLIP-style data used for contrastive training. This contrasts with the LLaVA-style autoregressive training, where long and highly descriptive captions (typically 200–500 tokens) are used to help the LVLM learn strong links between the vision and text domains, pay attention to fine-grained details, and develop strong reasoning and compositionality capabilities.

As noted earlier, directly using the long captions with the contrastive loss is ineffective, as, due to the high specificity of the long captions, the task is easy and nearly trivial to solve, with the loss going to $0$ in just a few hundred iterations. To address this, we propose to instead apply the next-token prediction loss over the long captions:
\vspace{-0.3cm}
\begin{equation}
    \mathcal{L}_{CE} = \sum_{i=1}^{L} \log p_\theta(u_i|\textbf{x}_v,\textbf{x}^{v}_p,\textbf{x}^{long}_{q,<i}),
    \vspace{-0.2cm}
\end{equation}
    where $L$ is the length of the long caption $\mathbf{x}^{long}_q$, $\mathbf{x}_v$ the input image, and $\mathbf{x}_{p}^v$ the prompt which prompts the model to describe the image in detail (e.g., ``Describe the image in detail''), and $p_\theta$ the next-token probability distribution learned by the model.

Intuitively, this formulation possesses multiple advantages: (1) It allows the model to learn from long captions, as predicting each and every token correctly is a challenging task (as opposed to applying the contrastive loss to long captions); (2) The decoding process encourages the condensation of information into the starting token used as a feature embedding; (3) It offers an avenue for retaining the generative capabilities of the model while strengthening its discriminative abilities.

\subsubsection{Overall training loss}
\label{ssec:method-total_loss}

As depicted in Fig.~\ref{fig:overall_framework}, we apply the next-token prediction loss over the long captions and the contrastive loss over the short ones in a unified manner. During training, the templates presented to the LVLM for the image and text modality take the following form:

\begin{minipage}{1.0\columnwidth}\vspace{2mm}\hspace{-7mm}    \centering
\begin{adjustbox}{width=1.0\columnwidth}
\begin{tcolorbox}
    \raggedright
    \small
      \vspace{-1mm}
     \textbf{Template for the image modality:}\\
     \texttt{USER: [Image Prompt]} \PredSty{\texttt{<image>}} \\\texttt{ASSISTANT:} \PredStyOrange{\texttt{<out\_token>}} \\\texttt{USER:} \texttt{Describe the image in detail.} \\\texttt{ASSISTANT:} \PredStyRed{\texttt{<long\_caption>}} \\ 
    \vspace{1mm}
    \textbf{Template for the text modality:}\\
      \texttt{USER: [Text Prompt]} \PredSty{\texttt{<short\_caption>}} \\ \texttt{ASSISTANT:} \PredStyOrange{\texttt{<out\_token>}}\\
     
     \vspace{-1mm}
\end{tcolorbox}
\end{adjustbox}
\vspace{1mm}
\end{minipage}
with the contrastive loss applied on the output representations {\color{myorange}\texttt{<out\_token>}} for the image modality and {\color{myorange}\texttt{<out\_token>}} for the text modality. Concomitantly, the next-token prediction loss is applied on the tokens of the {\color{myred}\texttt{<long\_caption>}}. Generally, the short caption must be sufficiently different from the long caption to prevent shortcuts during training, a property that naturally emerges in our case due to the difference in length and annotation procedure. Note that the distinction between long and short captions is made only during training. At test time, the model is used in discriminative mode as detailed in Sec.~\ref{ssec:background}.

\begin{figure*}[!ht]
    \centering
    \includegraphics[width=0.7\linewidth]{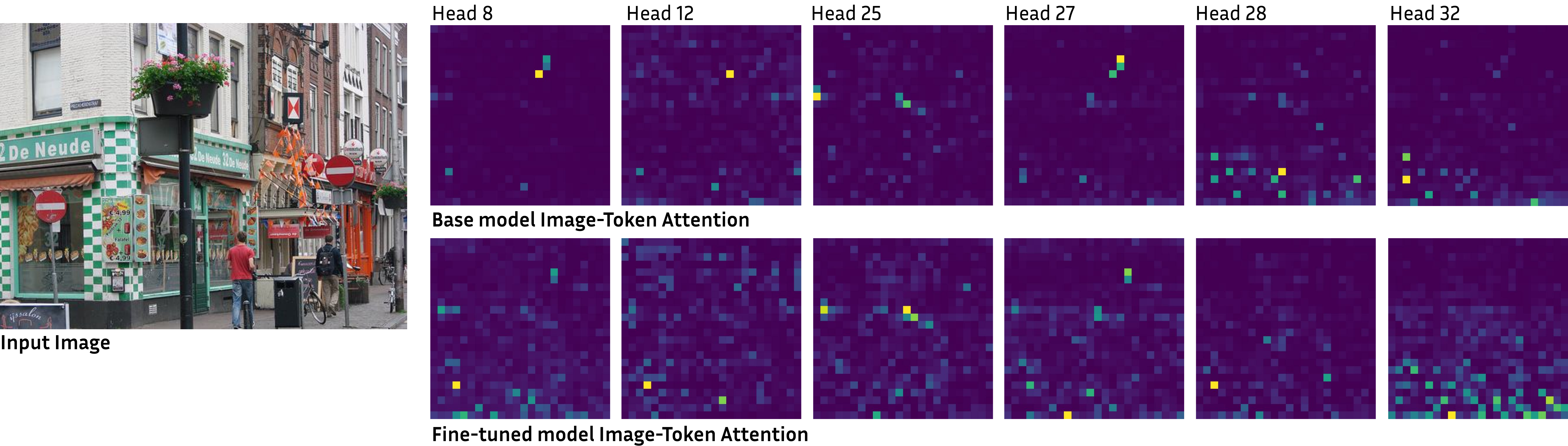}
    \vspace*{-0.2cm}
    \caption{\textbf{Attention map between the summary and vision tokens shown for a set of heads}. Notice that post-training, the attention maps densify. This behavioral change can be interpreted as follows: For generative tasks, at every step in the generation process, the model has the chance to look back at the vision tokens, selectively attending to the regions of interest at the current step. In contrast, in a discriminative setting, the model must compress all information present in the image within the summary token.}
    \label{fig:retrieval-prompts}
    \vspace{-0.25cm}
\end{figure*}

\subsection{Parameter-efficient adaptation}
\label{ssec:method-efficient}

As direct fine-tuning of the LVLM is costly, especially when maintaining a reasonably large batch size for contrastive learning, herein, we adopt parameter-efficient training with soft-prompting \textit{combined with} LoRA adapters, both trained under the same loss formulation of Sec.~\ref{ssec:method-generation-to-discrimination}.

\noindent \textbf{Soft prompting} was recently proposed as an efficient task-adaptation approach for both LLM~\cite{li2021prefix} and CLIP~\cite{zhou2022learning,bulat2023lasp} models, representing a direct departure from the prompt hand-crafting solution. Specifically, for a given input modality, \ie image and text, we define a set of $n$ modality($m$)-specific learnable vectors $[\mathbf{v}_1^m, \mathbf{v}_2^m, \cdots, \mathbf{v}_n^m]$, $\mathbf{v}_i^m \in \mathbb{R}^C$ with $C$ denoting the model's vocabulary embedding size. These vectors can be inserted across the input sequence to adjust the model's behavior. In practice, we opt to replace the tokens belonging to the hard prompts (\ie $\mathbf{x}_p^v$ and $\mathbf{x}_p^t$; see Sec.~\ref{ssec:background}) with the learnable vectors, initializing their values with the embeddings of the handcrafted ones. %

\noindent \textbf{Adapter fine-tuning:} While efficient, the representation power of the soft prompts is somewhat limited. Hence, following best practices, we also attach LoRA~\cite{hu2021lora} adapters to the linear layers located inside $\Phi_t$. Such adapters offer a multifold advantage: lower memory requirements, reduced potential of overfitting during training, and no additional compute requirements during inference. 

The model is fine-tuned using these components. Importantly, both have a positive impact on overall accuracy.

\subsection{How does the model's behavior change?}\label{ssec:model_beh}

Building upon the analysis from Sec.~\ref{ssec:background}, we show that our training approach elicits the following \textit{behavioral} changes:
(1) The attention map between the summary and vision tokens increases in density.
(2) Both the entropy of the output distribution of the summary token and the spread of the cumulative variance of the embeddings increase.

\noindent \textbf{The attention map densification}, as exemplified in Fig.~\ref{fig:retrieval-prompts}, shows that, for discriminative tasks, the model gathers evidence from all parts of the image in order to correctly encode the information therein. This is not needed for generation, as at every generation step, the model can ``peak back'' at the vision tokens and select the required information.

\noindent \textbf{Entropy and cumulative variance:} As shown in Fig.~\ref{fig:cumvar}, our approach results in models where the cumulative variance of the image and text embeddings is significantly more spread out, which translates into richer and better-aligned embeddings, capable of more accurately capturing fine-grained details. Additionally, the model maintains the diversity of output distribution at the summary token, \ie high entropy, as illustrated in Fig.~\ref{fig:decoded-tokens}.

%% file: sec/4_results.tex
\section{Experiments}
\label{sec:results}

\begin{table*}[!ht]
  \small
    \centering
      \caption{Zero-shot text-image retrieval accuracy on Flickr30K, COCO and nocaps.}
    \label{tab:zero-shot-retrieval}
    \vspace*{-0.3cm}
     \resizebox{\textwidth}{!}{
    \begin{tabular}{lcccccccccccc}
        \toprule
        & \multicolumn{6}{c}{\textbf{image} retrieval} & \multicolumn{6}{c}{\textbf{text} retrieval} \\
        \cmidrule(lr){2-7} \cmidrule(lr){8-13}
        \textbf{Method} & \multicolumn{2}{c}{Flickr30K} & \multicolumn{2}{c}{COCO} & \multicolumn{2}{c}{nocaps} & \multicolumn{2}{c}{Flickr30K} & \multicolumn{2}{c}{COCO} & \multicolumn{2}{c}{nocaps}  \\
        \cmidrule(lr){2-3} \cmidrule(lr){4-5} \cmidrule(lr){6-7} \cmidrule(lr){8-9} \cmidrule(lr){10-11} \cmidrule(lr){12-13}
        & R@1 & R@10 & R@1 & R@10 & R@1 & R@10 & R@1 & R@10  & R@1 & R@10 & R@1 & R@10\\
        \midrule
        CLIP (\texttt{ViT-L})~\cite{radford2021learning} & 67.3 & 93.3 & 37.0 & 71.5 & 48.6 & 85.7 & 87.2 & 99.4 & 58.1 & 87.8 & 70.0 & 96.2\\
        BLIP (\texttt{ViT-L})~\cite{li2022blip} & 70.0  & 95.2 & 48.4 & 83.2 & 62.3 & 93.4 & 75.5 & 97.7 & 63.5  & 92.5  & 72.1 & 97.7\\
        BLIP2 (\texttt{ViT-L})~\cite{li2023blip} & 74.5  & 97.2 & 50.0 & 86.1 & 63.0 & 93.8 & 86.1 & 99.4 & 63.0  & 93.1 & 74.4 & 98.3 \\
        OpenCLIP (\texttt{ViT-G/14})~\cite{schuhmann2022laion} & 77.8 & 96.9 & 48.8 & 81.5 & 63.7 & 93.2 & 91.5 & 99.6 & 66.3 & 91.8 & 81.0 & 98.7\\
        OpenCLIP (\texttt{ViT-BigG/14})~\cite{schuhmann2022laion} & 79.5 & 97.1 & 51.3 & 83.0 & 65.1 & 93.5 & 92.9 & 97.1 & 67.3 & 92.6 & 82.3 & 98.8 \\
        EVA-02-CLIP (\texttt{ViT-E/14+})~\cite{sun2023eva} & 78.8 & 96.8 & 51.1 & 82.7 & 64.5 & 92.9 & 93.9 & 99.8 & 68.8 & 92.8 & 83.0 & 98.9 \\
        EVA-CLIP (\texttt{8B})~\cite{sun2024eva} & 80.3 & 97.2 & 52.0 & 82.9 & 65.3 & 93.2 & 94.5 & 99.7 & 70.1 & 93.1 & 83.5 & 98.6 \\
        EVA-CLIP (\texttt{18B})~\cite{sun2024eva} & 83.3 & 97.9 & 55.6 & 85.2 & 69.3 & 94.8 & \textbf{95.3} & 99.8 & 72.8 & 94.2 & 85.6 & 99.2\\
        \midrule
        LLaVA-1.5-7B~\cite{liu2024improved} & 59.6 & 89.3 & 34.4 & 69.6 & 46.9 & 83.3 & 65.6 & 92.3 & 35.6 & 70.5 & 52.1 & 88.1 \\
        VLM2Vec (\texttt{Mistral-7B})~\cite{jiang2024vlm2vec} & 80.1 & 97.3 & 52.0 & 85.6 & 65.9 & 94.5 & 90.3 & 99.6 & 68.2 & 93.2 & 79.2  & 98.5 \\
        E5-V (\texttt{LLaVA-Next-8B})~\cite{jiang2024e5} & 79.5 & 97.6 & 52.0 & 84.7 &  65.9 & 94.3 & 88.2 & 99.4 & 62.0 & 89.7 & 74.9 & 98.3\\
        E5-V (\texttt{LLaVA-1.5-7B})~\cite{jiang2024e5} & 76.7 & 96.9 & 48.2 & 82.1 & 62.0 & 93.0 & 86.6 & 99.0 & 57.4 & 88.4 & 71.9 & 97.0 \\
        \rowcolor{Mycolor2} VladVA (Ours) (\texttt{LLaVA-1.5-7B}) & \textbf{85.0} & \textbf{98.5} & \textbf{59.0} & \textbf{88.6} & \textbf{72.3} & \textbf{96.5}  & 94.3 & \textbf{99.9} & \textbf{72.9} & \textbf{94.4} & \textbf{85.7} & \textbf{99.5}\\
        \bottomrule
    \end{tabular}
    }
    \vspace*{-0.4cm}
\end{table*}

We compare our approach with the current state-of-the-art on two tasks of interest in a zero-shot manner: image-text retrieval and compositionality/language understanding. 

\noindent \textbf{Models compared:} We compare with state-of-the-art models based on the two-towers (independent) approach, which is practical for retrieval purposes and also followed by our method. We cover a wide variety of settings: different models and model sizes, training data, training losses, etc.: CLIP (\texttt{ViT-L})~\cite{radford2021learning} $-$ the original CLIP trained with a contrastive loss on 400M image-text pairs; BLIP (\texttt{ViT-L})~\cite{li2022blip} $-$ trained on over 120M samples using contrastive, captioning and image-text matching losses; BLIP2 (\texttt{T5-XXL}) $-$ improved and scaled-up version of BLIP; OpenCLIP (\texttt{ViT-G/14})~\cite{schuhmann2022laion} $-$ scaled-up version of~\cite{radford2021learning} trained on 2B samples; OpenCLIP (\texttt{ViT-BigG/14})~\cite{schuhmann2022laion}, EVA-02-CLIP (\texttt{ViT-E/14+})~\cite{sun2023eva}, EVA-CLIP (\texttt{8B})~\cite{sun2024eva} and EVA-CLIP (\texttt{18B})~\cite{sun2024eva} $-$ large contrastively trained models, with up to 18B parameters, fine-tuned from vision encoders trained with Masked Image Modeling (MIM); E5-V (\texttt{LLaVA-Next-8B}) and E5-V (\texttt{LLaVA-1.5-7B}) $-$ LVLMs finetuned using a text-text contrastive loss. Depending on the task, we also include additional specialized baselines (\eg NegCLIP~\cite{yuksekgonul2022and} for compositionality).

\begin{table*}[!htbp]
  \centering
  \small
  \caption{Comparison with state-of-the-art on the SugarCrepe compositionality benchmark.}\label{tab:sota_eval_sc}
  \vspace*{-0.3cm}
  \resizebox{0.9\textwidth}{!}{
    \begin{tabular}{lcc ccccccc} 
    \toprule
      \multirow{ 2}{*}{Method} & Params & \multicolumn{3}{c}{ Replace}  &  \multicolumn{2}{c}{ Swap} & \multicolumn{2}{c}{ Add} \\ 
     \cmidrule(lr){3-5}\cmidrule(lr){6-7}\cmidrule(lr){8-9}
     
     & (B)  & Object & Attribute & Relation & Object & Attribute & Object & Attribute   \\ 
    \midrule
Human & -- & 100 & 99 & 97  & 99 & 100 & 99 & 99\\
\midrule
   CLIP (\texttt{ViT-L})~\cite{radford2021learning} & 0.43 & 94.1 & 79.2 & 65.2 & 60.2 & 62.3 & 78.3 & 71.5\\
   BLIP (\texttt{ViT-L})~\cite{li2022blip} & 0.23 & 96.5 & 81.7 & 69.1 & 66.6 & 76.8 & 92.0 & 85.1  \\
   BLIP2 (\texttt{ViT-L})~\cite{li2023blip} & 1.17 & 97.6 & 81.7 & 77.8 & 62.1 & 65.5 & 92.4 & 87.4 \\
   OpenCLIP (\texttt{ViT-G/14})~\cite{schuhmann2022laion} & 1.37 & 95.8& 85.0& 72.4& 63.0& 71.2& 91.5& 82.1\\
   OpenCLIP (\texttt{ViT-BigG/14})~\cite{schuhmann2022laion} & 2.54 & 96.6& 87.9& 74.9& 62.5& 75.2& 92.2& 84.5\\
   EVA-02-CLIP (\texttt{ViT-E/14+})~\cite{sun2023eva} & 5.04 & 97.1 & 88.5 & 74.2 & 67.3 & 74.1 & 91.8 & 83.9 \\
   EVA-CLIP (\texttt{8B})~\cite{sun2024eva} & 8.22 & 96.4 & 86.6 & 74.8 & 66.1 & 74.6 & 91.3 & 82.0 \\
   EVA-CLIP (\texttt{(18B})~\cite{sun2024eva} &  18.3 & 97.5 & 88.8 & 76.1 & 65.3 & 76.0 & 92.4 & 85.0 \\
\midrule
NegCLIP \cite{yuksekgonul2022and} & 0.15 & 92.7 & 85.9 & 76.5 & 75.2 & 75.4 & 88.8 & 82.8\\
\midrule
LLaVA-1.5-7B~\cite{liu2024improved} & 7.06 & 88.0 & 81.6 & 76.1 & 60.9 & 58.8 & 67.0 & 62.4 \\
VLM2Vec (\texttt{Mistral-7B})~\cite{jiang2024vlm2vec} &  7.30 & 97.2 & 89.0 & 81.7 & 62.9  &72.5 &94.7& 88.6 \\
E5-V (\texttt{LLaVA-Next-8B})~\cite{jiang2024e5} & 8.36 & 96.7 & 89.5 & 85.3 & 75.0 & 70.1 & 89.2 & 83.5 \\
E5-V (\texttt{LLaVA-1.5-7B})~\cite{jiang2024e5} & 7.06 & 95.8 & 86.6 & 81.6 & 62.9 & 64.0 & 93.5 & 88.0 \\
\rowcolor{Mycolor2} VladVA (Ours) (\texttt{LLaVA-1.5-7B}) & 7.06 & \textbf{98.1} & \textbf{92.1} & \textbf{86.8} & \textbf{79.0} & \textbf{82.9} & \textbf{95.2} & \textbf{95.8} \\
\bottomrule 
\end{tabular}
}
\vspace*{-0.25cm}
\end{table*}

\noindent \textbf{Training details:} We use a LLaVA-1.5 (7B)~\cite{liu2024visual} model due to its popularity and simplicity (for other models, see supp. material). For LoRA adapters, we set the rank and $\alpha$ to 16. The number of soft prompts is aligned to the length of the tokenized hand-crafted prompt.
Unless otherwise stated, we train the models for 7 epochs, using a batch size of 1024, a learning rate of $1e-4$, no weight decay, and AdamW~\cite{loshchilov2017decoupled} optimizer with default values for $\beta_1$ and $\beta_2$. During training, the learning rate is decayed according to a cosine scheduler~\cite{loshchilov2016sgdr}. Depending on the data configuration, we use up to 32 A100 GPUs. All our models and training procedures were implemented using PyTorch~\cite{paszke2019pytorch} and DeepSpeed~\cite{rasley2020deepspeed}.

We used the following training data: a 4M random subset of OpenImages~\cite{kuznetsova2020open}, CC3M ($\sim $2.8M images)~\cite{sharma2018conceptual}, and ShareGPT-4V~\cite{chen2023sharegpt4v}. As no captions are available for OpenImages, we automatically label them with 5 captions using BLIP2~\cite{li2023blip}. During training, only one caption is sampled at a time.
For longer captions, we directly use the ShareGPT-4V~\cite{chen2023sharegpt4v} data, which we extend with synthetic short captions produced by BLIP2 in order to enable the training procedure proposed in Sec.~\ref{ssec:method-total_loss}. Similarly, CC3M is automatically annotated with long captions using ShareGPT4-V~\cite{chen2023sharegpt4v}. %

\subsection{Zero-shot image-text retrieval}

We test our approach on the standard Flickr30k~\cite{young2014image}, MS-COCO~\cite{lin2014microsoft} and nocaps~\cite{agrawal2019nocaps} datasets, containing 1,000, 5,000 and 15,100 test samples respectively. For the latter, we simply average the results on the three partitions.

As shown in Tab.~\ref{tab:zero-shot-retrieval}, across all three datasets, our approach significantly surpasses the current state-of-the-art including models of similar size. It even outperforms the much bigger EVA-CLIP (\texttt{18B}) model (85.0\% vs. 83.3\%) on Flickr30k, (59.0\% vs. 55.6\%) on MS-COCO and (72.3\% vs. 69.3\%) on nocaps in terms of @R1 for image retrieval. Similarly, we outperform the LVLM-based E5-V model by 5.5\% on Flickr30k, 7\% on MS-COCO, and 6.4\% on nocaps.

\subsection{Image-text compositionality}

Herein, we focus our comparison on the currently most challenging test sets,  SugarCrepe~\cite{hsieh2024sugarcrepe} and SugarCrepe++~\cite{dumpala2024sugarcrepe++} (for Winoground~\cite{thrush2022winoground} please see supp. material). For SugarCrepe++, we are mostly interested in the Image-to-Text (ITT) setting since the Text-to-Text (TOT) one evaluates the language component of the methods only.

As Tabs.~\ref{tab:sota_eval_sc} and \ref{tab:sota_eval_sc_pp} show, our approach is the best in both SugarCrepe and SugarCrepe++ (ITT). On SugarCrepe, we outperform the 18B EVA-CLIP model on all categories, with particularly large gains on relation replacement (76.1 vs. 86.8), attribution adding (85.0 vs. 95.8), and object swap (65.3 vs. 79.0). The last case is particularly interesting as it directly measures the \textit{bag-of-words} behavior, showcasing  significant improvements offered by our method. Additionally, we outperform the E5-V variant based on the same LLaVA-1.5-7B model that we used, and the one based on the heavier LLaVA-Next-8B. A similar trend is observed on SugarCreppe++ where we outperform EVA-CLIP (\texttt{18B}) by up to 10.9\% (on object swap) and E5-V (ITT) in all but relation replacement. Thanks to its text-text training, E5-V surpasses our method for the TOT setting, but we note that their loss can be readily incorporated into our framework, leaving this for future work. %

\begin{table*}[!htbp]
\vspace{-0.1cm}
\caption{Comparison with state-of-the-art on the SugarCrepe++ compositionality benchmark.}
\label{tab:sota_eval_sc_pp}
\vspace*{-0.3cm}
  \centering
  \footnotesize
  \begin{tabular}{lccccccccccc}
  \toprule
   \multirow{ 2}{*}{Method} & Params & \multicolumn{2}{c}{Swap Object} & \multicolumn{2}{c}{Swap Attribute} & \multicolumn{2}{c}{Replace Object} & \multicolumn{2}{c}{Replace Attribute} & \multicolumn{2}{c}{Replace Relation} \\
  \cmidrule(lr){3-4} \cmidrule(lr){5-6} \cmidrule(lr){7-8} \cmidrule(lr){9-10} \cmidrule(lr){11-12}
   & (B) & \multicolumn{1}{c}{\ITT} & \multicolumn{1}{c}{\TOT} & \multicolumn{1}{c}{\ITT} & \multicolumn{1}{c}{\TOT} & \multicolumn{1}{c}{\ITT} & \multicolumn{1}{c}{\TOT} & \multicolumn{1}{c}{\ITT} & \multicolumn{1}{c}{\TOT} & \multicolumn{1}{c}{\ITT} & \multicolumn{1}{c}{\TOT} \\
   \midrule
   Human & -- & 100.00 & 96.7 & 96.7 & 93.3 & 100.00 & 97.00 & 100.00 & 98.3 & 100.00 & 96.7 \\
   \midrule
   CLIP (\texttt{ViT-L})~\cite{radford2021learning} & 0.43 & 46.0 & 14.5 & 44.5 & 28.7 & 92.0 & 81.3 & 68.8 & 56.3 & 53.4 & 39.1 \\
   BLIP (\texttt{ViT-L})~\cite{li2022blip} & 0.23 & 46.8 & 29.8 & 60.1 & 52.5 & 92.6 & 89.1 & 71.7 & 75.0 & 56.8 & 57.7  \\
   BLIP2 (\texttt{ViT-L})~\cite{li2023blip} & 1.17 & 37.9 & 39.5 & 51.9 & 55.4 & 94.8 & 96.9 & 73.2 & 86.5 & 65.1 & 69.6 \\
   OpenCLIP (\texttt{ViT-G/14})~\cite{schuhmann2022laion} & 1.37 & 40.7 & 27.4 & 54.2 & 49.6 & 93.1 & 89.4 & 72.5 & 73.1 & 57.6 & 51.4 \\
   OpenCLIP (\texttt{ViT-BigG/14})~\cite{schuhmann2022laion} & 2.54 & 48.8 & 28.2 & 57.7 & 52.4 & 94.2 & 90.5 & 76.4 & 72.6 & 59.4 & 53.6 \\
   EVA-02-CLIP (\texttt{ViT-E/14+})~\cite{sun2023eva} & 5.04 & 48.4 & 28.2 & 56.3 & 49.4 & 94.5 & 88.9 & 76.3 & 70.6 & 59.4 & 49.4 \\
   EVA-CLIP (\texttt{8B})~\cite{sun2024eva} & 8.22 & 43.6 & 25.4 & 55.2 & 46.9 & 93.7 & 85.8 & 73.4 & 67.9 & 59.7 & 49.2 \\
   EVA-CLIP (\texttt{18B})~\cite{sun2024eva} & 18.3 & 45.2 & 25.4 & 55.5 & 47.6 & 94.1 & 85.1 & 77.0 & 69.8 & 60.4 & 47.8 \\
   \midrule
  NegCLIP \cite{yuksekgonul2022and} & 0.15 & 55.3 & 34.7 & 58.0 & 56.5 & 89.5 & 94.5 & 69.4 & 76.3 & 52.3 & 51.6 \\
  CLIP-SVLC \cite{doveh2023teaching} & 0.15 & 43.0 & 18.9 & 48.4 & 34.6 & 80.9 & 91.6 & 57.0 & 66.9 & 47.3 & 51.3 \\
  BLIP-SGVL \cite{herzig2023incorporating} & 0.15 & 13.2 & \multicolumn{1}{c}{--} & 38.8 & \multicolumn{1}{c}{--} & 53.8 & \multicolumn{1}{c}{--} & 34.4 & \multicolumn{1}{c}{--} & 30.7 & \multicolumn{1}{c}{--} \\
  \midrule
  LLaVA-1.5-7B~\cite{liu2024improved} & 7.06 & 23.8 & 30.7 & 28.0 & 29.5 & 58.1 & 63.0 & 46.8 & 58.1 & 52.3 & 63.4 \\
  VLM2Vec (\texttt{Mistral-7B})~\cite{jiang2024vlm2vec} & 7.30 & 40.7 & 39.9 & 48.1 & 50.0 & 94.6 & 96.9 & 77.0 & 85.6 & 67.9 & 70.7 \\
  E5-V (\texttt{LLaVA-Next-8B})~\cite{jiang2024e5} & 8.36 & 50.8 & \textbf{48.4} & 49.7 & 56.9 & 93.1 & \textbf{97.6}  & 76.1 & \textbf{87.1} & \textbf{74.7} & \textbf{84.4}\\
  E5-V (\texttt{LLaVA-1.5-7B})~\cite{jiang2024e5} & 7.06 & 39.5 & 42.3 & 40.7 & 48.5 & 89.7 & 94.6  & 71.7 & 86.4 & 72.0 & 81.5\\
  \rowcolor{Mycolor2} VladVA (Ours) (\texttt{LLaVA-1.5-7B}) & 7.06 & \textbf{56.1} & 36.7 & \textbf{63.0} & \textbf{62.5} & \textbf{95.0}  & 93.0 & \textbf{78.2} & 82.3 & 71.1 & 66.3\\
  \bottomrule
  \end{tabular}
  \vspace*{-0.25cm}
\end{table*}

%% file: sec/5_ablation_and_analysis.tex
\section{Ablation studies}
\label{sec:ablations}

\subsection{Impact of method's components}

We quantify the impact of the proposed method's components by training on a smaller 1M subset, reporting results on SugarCrepe (averaged over each category) and on Flickr30k (R@1 for T2I and I2T). 

\noindent \textbf{Impact of adaptation components:} We start by measuring the impact of the efficient adaptation strategy based on soft prompting and adapter-finetuning. For simplicity, we ablate this by training using only the contrastive loss. As the results from Tab.~\ref{tab:ablation_components} show, both components, individually and jointly, provide notable gains on top of the original LLaVA-1.5-7B model (i.e. the case of no adaptation). 

While LoRA fine-tuning performs better than soft-prompting (due to its bigger capacity), the latter alone performs surprisingly well. To understand why, we analyze the changes the soft prompts undergo by finding the closest embedding in the LLM's vocabulary. This results in the following decoded sentences: ``\texttt{</s> '<Summarize the provided image in one word:/ \$[}'' and, ``\texttt{ $\omega$aSummarize the provided text in one word:$-$}''. The two sentences remain unchanged semantically, with the only characters changed being the ones at the start and the end of the prompt. Intuitively, this allows the model to mark/specialize the token that should gather the visual or textual evidence for discriminative tasks.

\noindent \textbf{Impact of AR loss:} We measure the impact of the proposed autoregressive loss on long captions from Sec.~\ref{ssec:method-ar}. As Tab.~\ref{tab:ablation_components} shows, the AR loss adds a notable performance boost across all datasets tested. Finally, we note that using the long captions in isolation, without the proposed training strategy and loss, does not result in measurable gains.

\begin{table}[t]
  \centering
  \small
    \caption{Impact of adaptation components and AR loss. All models are trained on 1M samples.}\label{tab:ablation_components}
    \vspace{-0.3cm}
  \resizebox{.48\textwidth}{!}{
    \begin{tabular}{lc ccccccc} 
    \toprule
         \multirow{ 2}{*}{Method}  & AR & \multicolumn{3}{c}{ SugarCrepe}  &  \multicolumn{2}{c}{ Flickr30k} \\ 
     \cmidrule(lr){3-5}\cmidrule(lr){6-7} 
     & loss &  Replace & Swap & Add & T2I & I2T\\
    \midrule 
LLaVA-1.5-7B   & \xmark & 81.9 & 59.8 & 64.7 & 59.6 & 65.6 \\
\midrule
$+$ soft-prom. & \xmark & 86.4 & 66.9 & 89.3 & 76.7 & 91.7\\
$+$ adapter-ft. & \xmark & 87.0 & 69.8 & 88.8 & 79.1 & 91.4\\
$+$ adapter-ft. + soft-prom. & \xmark & 87.1 & 72.0 & 88.6 & 79.6 & \textbf{92.9}\\
$+$ adapter-ft. + soft-prom. &  \checkmark &\textbf{89.5} & \textbf{75.5} & \textbf{89.5} & \textbf{80.6} & 91.8\\
\bottomrule 
\end{tabular}
}
\vspace{-0.2cm}
\end{table}

\begin{table}[t]
  \centering
  \small
  \caption{Impact of training data size.}\label{tab:ablations_data}
  \vspace{-0.3cm}
  \resizebox{.48\textwidth}{!}{
    \begin{tabular}{lc ccccccc} 
    \toprule
     \multirow{ 2}{*}{Training data}  & \multicolumn{3}{c}{ SugarCrepe}  &  \multicolumn{2}{c}{ Flickr30k} \\ 
     \cmidrule(lr){2-4}\cmidrule(lr){5-6} 
      &  Replace & Swap & Add & T2I & I2T\\
    \midrule
LLava-1.5-7B (0M)   & 81.9 & 59.8 & 64.7 & 59.6 & 65.6 \\
\midrule
OpenImages (1M) & 87.1 & 72.0 & 88.6 & 79.6 & 92.9 \\
OpenImages (4M) & 88.2 & 79.6 & 89.1 & 82.3 & 93.1\\
~~~$+$ ShareGPT-4V (1.3M) & 91.0 & 80.3 & 92.4 & 83.1 & 94.0 \\
~~~~~~$+$ CC3M (2.8M) & \textbf{92.3} & \textbf{80.9} & \textbf{95.5} & \textbf{85.0} & \textbf{94.3} \\
\bottomrule 
\end{tabular}
\vspace{-0.3cm}
}
\end{table}

\subsection{Impact of training dataset size}

Although at a relatively small scale (training is expensive due to the LVLM), herein, we aim to examine whether scaling the dataset size benefits the proposed discriminative adaptation of LVLMs. Specifically, we scale our dataset size from 1M to 8.1M samples. As Tab.~\ref{tab:ablations_data} shows, we obtain steady gains across all metrics, with no signs of immediate saturation. This suggests that some potential is still left untapped, and further scaling could result in extra gains.

%% file: sec/6_conclusions.tex
\section{Conclusions}

We introduced a new framework for adapting a \textit{generative} LVLM into a \textit{discriminative} model, unlocking its innate capability for powerful image-text discrimination and enhanced language understanding. 
Our framework uses both short and long captions for training the LVLM with contrastive and next-token prediction losses respectively. We also presented a parameter-efficient adaptation method using a combination of soft prompting and LoRA adapters. Finally, we showed that our approach results in significant improvements over state-of-the-art models of similar size for image-text retrieval and compositionality benchmarks.

%% file: sec/X_suppl.tex
\clearpage
\setcounter{page}{1}
\maketitlesupplementary

\section{Results for additional model sizes and architectures}

\begin{table*}[!hb]
  \small
    \centering
      \caption{Zero-shot text-image retrieval accuracy on Flickr30K, COCO and nocaps.}
    \label{tab:zero-shot-retrieval}
     \resizebox{\textwidth}{!}{
    \begin{tabular}{lcccccccccccc}
        \toprule
        & \multicolumn{6}{c}{\textbf{image} retrieval} & \multicolumn{6}{c}{\textbf{text} retrieval} \\
        \cmidrule(lr){2-7} \cmidrule(lr){8-13}
        \textbf{Method} & \multicolumn{2}{c}{Flickr30K} & \multicolumn{2}{c}{COCO} & \multicolumn{2}{c}{nocaps} & \multicolumn{2}{c}{Flickr30K} & \multicolumn{2}{c}{COCO} & \multicolumn{2}{c}{nocaps}  \\
        \cmidrule(lr){2-3} \cmidrule(lr){4-5} \cmidrule(lr){6-7} \cmidrule(lr){8-9} \cmidrule(lr){10-11} \cmidrule(lr){12-13}
        & R@1 & R@10 & R@1 & R@10 & R@1 & R@10 & R@1 & R@10  & R@1 & R@10 & R@1 & R@10\\
        \midrule
        Qwen2-VL-2B~\cite{wang2024qwen2} & 54.1 & 86.0 & 32.4 & 68.2 & 41.2 & 80.1 & 59.6 & 89.2 & 35.3 & 71.8 & 54.0 & 90.3 \\
        \rowcolor{Mycolor2} VladVA (Ours) (\texttt{Qwen2-VL-2B}) & \textbf{80.4} & \textbf{97.3} & \textbf{52.5} & \textbf{84.4} & \textbf{68.3} & \textbf{94.9} & \textbf{93.7} & \textbf{99.9} & \textbf{71.9} & \textbf{93.9} & \textbf{86.0} & \textbf{99.4}  \\
        \midrule
        LLaVA-1.5-7B~\cite{liu2024improved} & 59.6 & 89.3 & 34.4 & 69.6 & 46.9 & 83.3 & 65.6 & 92.3 & 35.6 & 70.5 & 52.1 & 88.1 \\
        \rowcolor{Mycolor2} VladVA (Ours) (\texttt{LLaVA-1.5-7B}) & \textbf{85.0} & \textbf{98.5} & \textbf{59.0} & \textbf{88.6} & \textbf{72.3} & \textbf{96.5}  & \textbf{94.3} & \textbf{99.9} & \textbf{72.9} & \textbf{94.4} & \textbf{85.7} & \textbf{99.5}\\
        \midrule
        LLaVA-1.5-13B~\cite{liu2024improved} & 61.7 & 90.4 & 37.9 & 74.1 & 48.4 & 85.0 & 66.9 & 93.6 & 35.3 & 71.0 & 48.0 & 87.9 \\
        \rowcolor{Mycolor2} VladVA (Ours) (\texttt{LLaVA-1.5-13B}) & \textbf{85.6} & \textbf{98.6} & \textbf{58.2} & \textbf{88.4} & \textbf{74.0} & \textbf{96.6} & \textbf{94.5} & \textbf{99.8} & \textbf{75.0} & \textbf{95.6} & \textbf{85.4} & \textbf{99.6} \\

        \bottomrule
    \end{tabular}
    }
    \vspace*{-0.2cm}
\end{table*}

\begin{table*}[!hb]
  \centering
  \small
  \caption{Zero-shot results on SugarCrepe compositionality benchmark.}\label{tab:sota_eval_sc_supmat}
  \resizebox{0.9\textwidth}{!}{
    \begin{tabular}{lcc ccccccc} 
    \toprule
      \multirow{ 2}{*}{Method} & Params & \multicolumn{3}{c}{ Replace}  &  \multicolumn{2}{c}{ Swap} & \multicolumn{2}{c}{ Add} \\ 
     \cmidrule(lr){3-5}\cmidrule(lr){6-7}\cmidrule(lr){8-9}
     
     & (B)  & Object & Attribute & Relation & Object & Attribute & Object & Attribute   \\ 
\midrule
Qwen2-VL-2B~\cite{wang2024qwen2} & 2.21 & 89.9 & 72.0 & 75.0 & 56.1 & 56.1 & 73.2 & 70.1 \\
\rowcolor{Mycolor2} VladVA (Ours) (\texttt{Qwen2-VL-2B}) & 2.21 & \textbf{97.9} & \textbf{89.7} & \textbf{81.5} & \textbf{76.5} & \textbf{82.6} & \textbf{93.6} & \textbf{95.4} \\
\midrule
LLaVA-1.5-7B~\cite{liu2024improved} & 7.06 & 88.0 & 81.6 & 76.1 & 60.9 & 58.8 & 67.0 & 62.4 \\
\rowcolor{Mycolor2} VladVA (Ours) (\texttt{LLaVA-1.5-7B}) & 7.06 & \textbf{98.1} & \textbf{92.1} & \textbf{86.8} & \textbf{79.0} & \textbf{82.9} & \textbf{95.2} & \textbf{95.8} \\
\midrule
LLaVA-1.5-13B~\cite{liu2024improved} & 13.35 & 90.0 & 80.6 & 76.3 & 71.8 & 61.9 & 69.3 & 59.1 \\
\rowcolor{Mycolor2} VladVA (Ours) (\texttt{LLaVA-1.5-13B}) & 13.35 & \textbf{98.1} & \textbf{93.9} & \textbf{89.8} & \textbf{81.1} & \textbf{86.0} & \textbf{95.2} & \textbf{97.0} \\

\bottomrule 
\end{tabular}
}
\end{table*}

\begin{table*}[!hb]
\caption{Zero-shot results on the SugarCrepe++ compositionality benchmark.}
\label{tab:sota_eval_sc_pp_supmat}
  \centering
  \footnotesize
  \begin{tabular}{lccccccccccc}
  \toprule
   \multirow{ 2}{*}{Method} & Params & \multicolumn{2}{c}{Swap Object} & \multicolumn{2}{c}{Swap Attribute} & \multicolumn{2}{c}{Replace Object} & \multicolumn{2}{c}{Replace Attribute} & \multicolumn{2}{c}{Replace Relation} \\
  \cmidrule(lr){3-4} \cmidrule(lr){5-6} \cmidrule(lr){7-8} \cmidrule(lr){9-10} \cmidrule(lr){11-12}
   & (B) & \multicolumn{1}{c}{\ITT} & \multicolumn{1}{c}{\TOT} & \multicolumn{1}{c}{\ITT} & \multicolumn{1}{c}{\TOT} & \multicolumn{1}{c}{\ITT} & \multicolumn{1}{c}{\TOT} & \multicolumn{1}{c}{\ITT} & \multicolumn{1}{c}{\TOT} & \multicolumn{1}{c}{\ITT} & \multicolumn{1}{c}{\TOT} \\
   \midrule
    Qwen2-VL-2B~\cite{wang2024qwen2} & 2.21 &  32.7 & 27.8 & 30.5 & 25.3 & 73.6 & 65.9 & 46.8 & 43.0  & 57.6 & \textbf{58.3} \\
    \rowcolor{Mycolor2} \textbf{VladVA (Ours)} (\texttt{Qwen2-VL-2B}) & 2.21 & \textbf{50.8} & \textbf{33.5} & \textbf{60.4} & \textbf{48.2} & \textbf{93.7}  & \textbf{93.8} & \textbf{74.8} & \textbf{77.5} & \textbf{63.6} & 57.4\\
  \midrule
  LLaVA-1.5-7B~\cite{liu2024improved} & 7.06 & 23.8 & 30.7 & 28.0 & 29.5 & 58.1 & 63.0 & 46.8 & 58.1 & 52.3 & 63.4 \\
  \rowcolor{Mycolor2} \textbf{VladVA (Ours)} (\texttt{LLaVA-1.5-7B}) & 7.06 & \textbf{56.1} & 36.7 & \textbf{63.0} & \textbf{62.5} & \textbf{95.0}  & 93.0 & \textbf{78.2} & 82.3 & 71.1 & 66.3\\
  \midrule
  LLaVA-1.5-13B~\cite{liu2024improved} & 13.35 & 35.5 & 32.3 & 30.2 & 32.4 & 68.7 & 66.8 & 44.8 & 43.1 & 52.3 & 55.6 \\
    \rowcolor{Mycolor2} \textbf{VladVA (Ours)} (\texttt{LLaVA-1.5-13B}) & 13.35 & \textbf{55.2} & \textbf{38.3} & \textbf{65.6} & \textbf{60.6} & \textbf{94.5}  & \textbf{92.5} & \textbf{80.7} & \textbf{81.1} & \textbf{73.2} & \textbf{66.4}\\
  
  \bottomrule
  \end{tabular}
\end{table*}

To further showcase the generalizability of our approach, herein we report results on two additional models: LLaVA-1.5-13B~\cite{liu2024improved} and Qwen2-VL-2B~\cite{wang2024qwen2}. The 1st is a scaled-up version of the LLaVA-1.5-7B~\cite{liu2024improved} used in the main manuscript and tests the scalability of our approach with size. The second follows a different architecture and training procedure and has ``only'' 2B parameters, testing both generalizations to different architectures and finetuning in a lower-parameters regime. As the results from Tables~\ref{tab:zero-shot-retrieval},~\ref{tab:sota_eval_sc_supmat}~\ref{tab:sota_eval_sc_pp_supmat} and~\ref{tab:sota_eval_winoground} show, on all 6 datasets (\ie Flickr, coco, nocaps, SugarCrepe, SugarCrepe++ and Winoground) for both retrieval and compositionality, in all cases we significantly improve upon the original zero-shot model performance, showing good scalability with size in both directions, \ie for smaller and bigger models.

\section{Compositionality evaluation on Winoground}

In addition to the results from the main paper, herein, we report results on Winoground~\cite{thrush2022winoground}, a curated dataset consisting of 400 images with difficult/unusual scenarios that go beyond compositionality and largely act as a natural adversarial set~\cite{diwan2022winoground,yuksekgonul2022and}. As the results from Table~\ref{tab:sota_eval_winoground} show, our approach matches and outperforms prior models, including the large 18B EVA-CLIP model (17.5 vs. 15.0, 40.5 vs. 35.8 and 12.8 vs. 10.5, for image, text and respectively group set).

\begin{table}[!h]
  \small
    \centering
      \caption{Comparison with state-of-the-art on the Winoground compositionality benchmark.}
    \label{tab:sota_eval_winoground}
    \begin{tabular}{lcccccccc}
        \toprule
        Model & Image & Text & Group\\
        \midrule
        CLIP (\texttt{ViT-B})~\cite{radford2021learning} & 10.5 & 25.0 & 7.3  \\
        CLIP (\texttt{ViT-L})~\cite{radford2021learning} & 12.3 & 27.5 & 8.3 \\
        BLIP (\texttt{ViT-L})~\cite{li2022blip} & 10.0  & 30.5 & 7.8 \\
        BLIP2 (\texttt{ViT-L})~\cite{li2023blip} & 10.5 & 29.5 & 8.5 \\
        OpenCLIP (\texttt{ViT-G/14})~\cite{schuhmann2022laion} & 12.8 & 32.0 & 9.3 \\
        OpenCLIP (\texttt{ViT-BigG/14})~\cite{schuhmann2022laion} & 15.5 & 35.5 & 12.0 \\
        EVA-02-CLIP (\texttt{ViT-E/14+})~\cite{sun2023eva} & 14.0 & 33.8 & 10.8 \\
        EVA-CLIP (\texttt{8B})~\cite{sun2024eva} & 14.8 & 36.5 & 10.3 \\
        EVA-CLIP (\texttt{18B})~\cite{sun2024eva} & 15.0 & 35.8 & 10.5\\
        \midrule
        NegCLIP~\cite{yuksekgonul2022and} & 10.5 & 29.5 & 8.0 \\
        \midrule
        LLaVA-1.5-7B~\cite{liu2024improved} & 11.3 & 18.5 & 6.5 \\
        E5-V (\texttt{LLaVA-Next-8B})~\cite{jiang2024e5} & 14.8 & 32.3 & 11.3 \\
        E5-V (\texttt{LLaVA-1.5-7B})~\cite{jiang2024e5} & 17.4 & 31.3 & 10.5 \\
        \rowcolor{Mycolor2} VladVA (Ours) (\texttt{LLaVA-1.5-7B}) & \textbf{17.5} & \textbf{40.5} & \textbf{12.8} \\
        \bottomrule
    \end{tabular}
\end{table}

\section{Zero-shot image recognition on ImageNet}

\begin{table}[!h]
  \small
    \centering
      \caption{Zero-shot image recognition results on ImageNet dataset in terms of Top-1 and Top-5 (\%) accuracy.}
    \label{tab:sota_eval_imagenet}
     \begin{adjustbox}{width=\columnwidth}
    \begin{tabular}{l*{5}{>{\rowfonttype}c}<{\rowfont{}}}
        \toprule
        Model & Data. size & Top-1 & Top-5 \\
        \midrule
        \rowfont{\color{gray}} CLIP (\texttt{ViT-B})~\cite{radford2021learning} & 400M & 68.4 & 91.9  \\
        \rowfont{\color{gray}} CLIP (\texttt{ViT-L})~\cite{radford2021learning} & 400M & 74.0 & 94.0 \\
        \rowfont{\color{gray}} EVA-CLIP (\texttt{18B})~\cite{sun2024eva} & 2.7B & 83.5 & 97.2 \\
        \midrule
         \rowfont{\color{black}} CLIP (\texttt{ViT-B})~\cite{radford2021learning} & 15M & 32.8 & - \\
        HiDeCLIP (\texttt{ViT-B})~\cite{radford2021learning} & 15M & 45.9 & - \\
        FFF (\texttt{ViT-B})~\cite{bulat2024fff} & 15M & 51.1 & -\\
        \midrule        
        \rowfont{\color{black}} BLIP (\texttt{ViT-L})~\cite{li2022blip} & 129M & 54.2 & 81.5 \\
        BLIP2 (\texttt{ViT-L})~\cite{li2023blip} & 129M & 46.7 & 74.2 \\
        \midrule
        LLaVA-Next-8B~\cite{li2024llavanext} & 0M & 45.8 & 74.6 \\
        E5-V~\cite{jiang2024e5} (\texttt{LLaVA-Next-8B}) & 0M & 48.2 & 76.6 \\
        \midrule 
        LLaVA-1.5-7B~\cite{liu2024improved} & 0M & 42.0 & 74.6 \\
        \rowcolor{Mycolor2} VladVA (Ours) (\texttt{LLaVA-1.5-7B}) & 8.1M & 63.7 & 88.3 \\
        Qwen2-VL-2B~\cite{wang2024qwen2} & 0M & 54.7 & 79.4 \\
        \rowcolor{Mycolor2} VladVA (Ours) (\texttt{Qwen2-VL-2B}) & 8.1M & 70.6 & 91.1 \\
        \bottomrule
    \end{tabular}
     \end{adjustbox}
\end{table}

From an evaluation point of view, the main focus of this work is on improved zero-shot retrieval and, more generally, improved vision-language compositional ability. We focus on these tasks, as they require stronger (vision-)language understanding abilities, which we show an LVLM can offer under appropriate training regimes. As a study case, herein, for completeness, we also measure the zero-shot ability of the model for image recognition on ImageNet~\cite{deng2009imagenet}. As the results from Table~\ref{tab:sota_eval_imagenet} show, our approach significantly improves upon the zero-shot LVLM we start from (54.7 vs 70.6\%). In comparison, E5-V approach only offers modest performance gains (45.8 vs 48.2\%) and has notably lower performance than our approach (48.2 vs 70.6\%) despite using a bigger model. While significantly improving upon the model we start from, the low data regime we train our model in (only 8.1M samples) limits its overall performance, with contrastive models trained on billion samples performing better. This is expected as the image recognition ability of a model, especially on the highly specific categories of ImageNet, will depend on how often (if at all) they are seen in the training set. This is especially significant given that many of the datasets used for contrastive learning are filtered based on the ImageNet classes~\cite{radford2021learning}. In lower data regimes, comparable with ours, we can observe that our approach produces notably better results (\eg 51.1\% for FFF~\cite{bulat2024fff}, trained on 15M samples vs 70.6\% for ours). Finally, when comparing it with other models focusing on retrieval (\ie BLIP and BLIP2) our approach outperforms either of them by more than 15\% in absolute terms despite the fact that these models were trained on 129M samples. All in all, we outperform all models trained in comparable settings, showing promising initial results in this direction too.

\section{Which layer to choose the token from?} 

In the main paper, we've used the last token of the last layer as the summary, discriminative token. Intuitively, by selecting the last layer, we maximize the amount of parameters we can adapt, and hence adaptation plasticity. 
However, herein, for completeness, we report results for different layer IDs in Table~\ref{tab:eval_layer_id}. The results show that the last 3-4 layers have comparable performance, performance that degrades as we select earlier layers.

\begin{table}[!h]
  \small
    \centering
      \caption{Performance change when using different layer IDs, reported on SugarCrepe (averaged) and Flickr30k (I2T).}
    \label{tab:eval_layer_id}

\begin{tabular}{l|cccccc}
\toprule
   Dataset/Layer & 32 (last)  & 31  & 28 & 24 & 20 & 16 \\
   \midrule
   Flickr30K  & 0 & +0.3 & -0.2 & -1.1 & -13.5 & -59.0 \\
   SugarCrepe: & 0 & +0.9 & +0.4 & -0.1 & -8.1 & -20.0 \\
     \bottomrule
\end{tabular}
\end{table}

\section{Qualitative text generation examples post discriminative adaptation}

\begin{figure*}[!ht]
    \centering
    \includegraphics[width=0.9\linewidth]{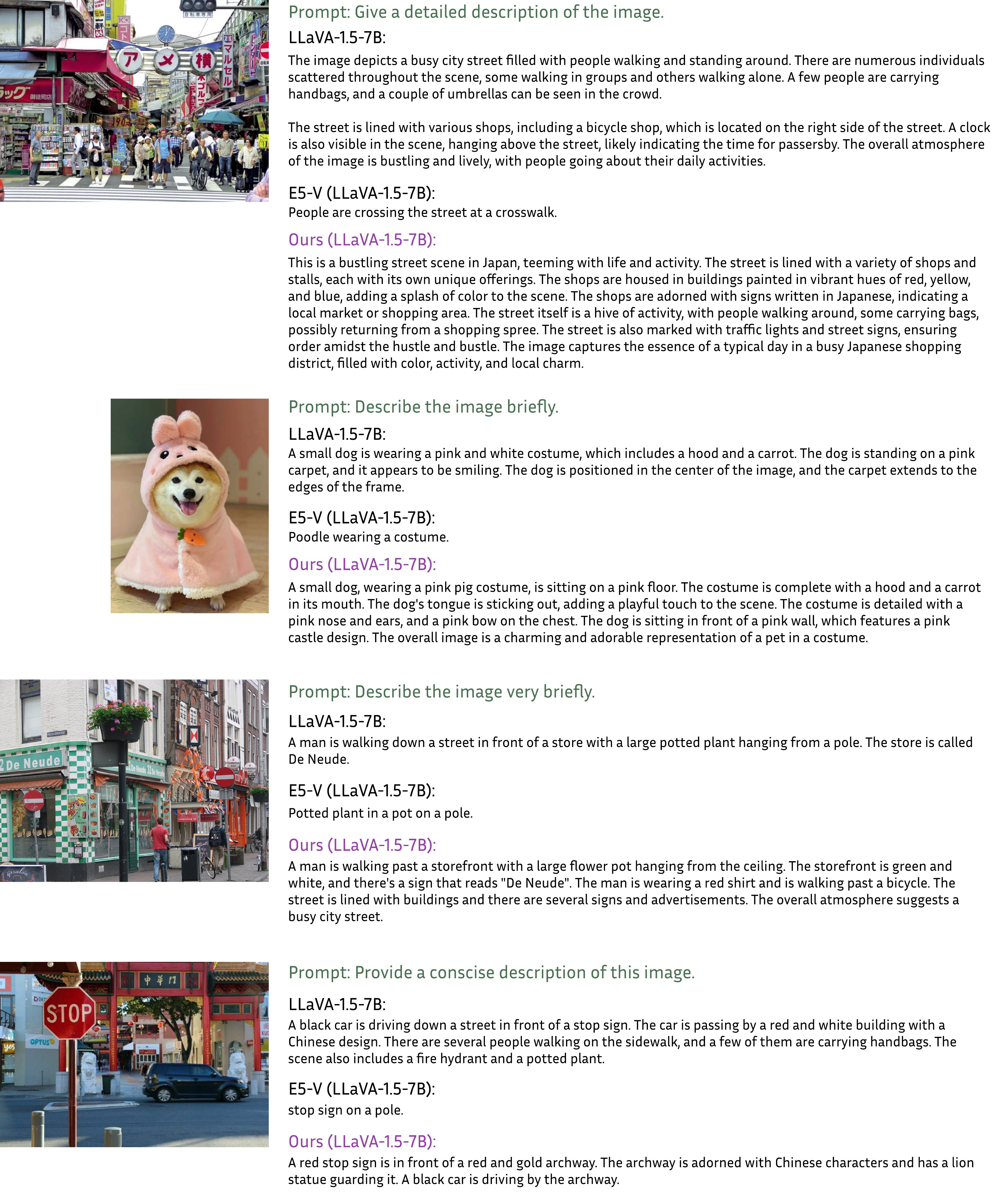}
    \caption{\textbf{Qualitative comparison on image captioning} of the base LLaVA-1.5-7B model and its fine-tuned versions using both E5-V~\cite{jiang2024e5} and our proposed method. We show that with our method, the LLaVA-1.5-7B better retains its captioning capabilities, while E5-V fine-tuning appears to result in less informative captions. }
    \label{fig:gen-captioning}
    \vspace{-0.4cm}
\end{figure*}

Our main objective is to convert generative LVLMs into discriminative ones, hence  the proposed approach is designed from the perspective of maximizing the discriminative abilities of the model. Still, it may be interesting to qualitatively see how our model, and the closest relevant approach E5-V behave. We note, that in principle both our approach and E5-V use LoRAs adapters, hence it is easy to switch between the discriminative and the generative mode without compromising either, by enabling or disabling the adapters. That being said, herein we present some qualitative examples post-training, so we can see the direct effect the training has on the model. As the results from Fig.\ref{fig:gen-captioning} show, generally, our approach better retains the generative capabilities of the model post-training, producing fine-grained captions, similar with the original ones. In contrast, E5-V appears to predominantly produce only very-shot, not-descriptive outputs.

\clearpage